\documentclass[runningheads,a4paper]{llncs}

\usepackage{amssymb}
\usepackage{graphicx}

\usepackage{url}
\urldef{\mailsa}\path|{alexander.hagg, alexander.asteroth}@h-brs.de|    
\urldef{\mailsb}\path|t.h.w.baeck@liacs.leidenuniv.nl|    
\newcommand{\keywords}[1]{\par\addvspace\baselineskip
\noindent\keywordname\enspace\ignorespaces#1}

\usepackage{makeidx}  
\usepackage{amsmath}
\usepackage{amsfonts}
\usepackage{placeins}
\usepackage[caption = true]{subfig}

\usepackage{tipa}
\usepackage{multicol}
\usepackage{gensymb}
\usepackage{algorithm,algorithmicx,algpseudocode}
\usepackage[misc,geometry]{ifsym} 
\usepackage[compatibility=false]{caption}

\begin{document}
	
\mainmatter
\title{Prototype Discovery using Quality-Diversity}

\author{Alexander Hagg\inst{1,2}\textsuperscript{(\Letter)} \and Alexander Asteroth\inst{1} \and Thomas B\"ack\inst{2}}
\authorrunning{Alexander Hagg\and Alexander Asteroth\and Thomas B\"ack}

\institute{Bonn-Rhein-Sieg University of Applied Sciences, Sankt Augustin, Germany\\\mailsa \and Leiden Institute of Advanced Computer Science, Leiden University, Leiden, The Netherlands\\\mailsb}

\maketitle

\begin{abstract}
	An iterative computer-aided ideation procedure is introduced, building on recent quality-diversity algorithms, which search for diverse as well as high-performing solutions. Dimensionality reduction is used to define a similarity space, in which solutions are clustered into classes. These classes are represented by prototypes, which are presented to the user for selection. In the next iteration, quality-diversity focuses on searching within the selected class. A quantitative analysis is performed on a 2D airfoil, and a more complex 3D side view mirror domain shows how computer-aided ideation can help to enhance engineers' intuition while allowing their design decisions to influence the design process.
\end{abstract}

\section{Introduction}
Conceptual engineering design is an iterative process~\cite{Flager2007}. Under the paradigm commonly called ideation~\cite{Bradner2014} a design problem is defined, the design space explored, candidate solutions evaluated, and finally design decisions are taken, which put constraints onto the next design iteration.

In a 2014 interview study by Bradner~\cite{Bradner2014} on the real-world usage of automation in design optimization, ``participants reported consulting Pareto plots iteratively in the conceptual design phase to rapidly identify and select interesting solutions''. This process of \textit{a posteriori articulation of preference}~\cite{Hwang1979} is described by the ``design by shopping'' paradigm~\cite{Balling1999,Stump2003}. A Pareto front of optima is created by a multi-objective optimization algorithm, after which engineers choose a solution to their liking. That participants used optimization algorithms to develop preliminary solutions to solve a problem surprised the interviewers.

Design optimization has been applied to multi-modal problems, using niching and crowding to enforce diversity in evolutionary optimization algorithms~\cite{Shir2005,Singh2006}. 
For optimization algorithms to operate effectively in cases where evaluation of designs is computationally expensive, surrogate assistance is applied, using predictive models that replace most of the evaluations~\cite{Jin2011}. Recently introduced quality-diversity (QD) algorithms, like NSLC~\cite{Lehman2011} and MAP-Elites~\cite{Mouret2015}, are evolutionary algorithms capable of producing a large array of solutions constituting different behaviors or design features. Surrogate assistance was introduced for QD algorithms~\cite{Gaier2017} as well. It enables finding thousands of designs, using orders of magnitude fewer evaluations than running MAP-Elites without a surrogate. However, this large number of solutions can hinder the engineer's ability to select interesting designs. 

As automated diversity gives too many solutions, their more concise presentation makes QD more useful to designers. In this paper we apply the design by shopping paradigm to QD, assisting design decisions by representing similar solutions succinctly with a representative solution using prototype theory. Therein, objects are part of the same class based on resemblance. Wittgenstein~\cite{Wittgenstein1953} questioned whether classes can even be rigidly limited, implying that there is such a thing as a distance to a class. Rosch later introduced prototype theory~\cite{Rosch1975}, stating that natural classes consist of a prototype, the best representative of its class, and non-prototypical examples, which can be ranked in terms of distance to the prototype. However, while feature diversity is enforced, surrogate-assisted QD uses no metric for genotypic similarity in terms of the actual design space.

\begin{figure}
	\centering
	\includegraphics{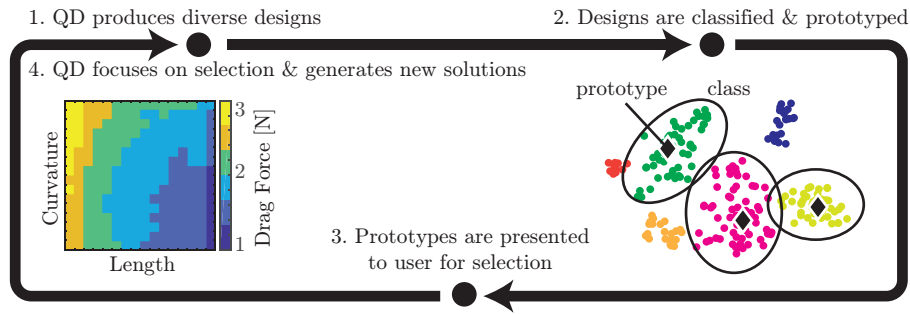}
	\caption{Computer-aided ideation loop. Step 1: QD algorithm is used to discover diverse optimal solutions. Step 2: similar solutions are grouped into classes. Step 3: prototypes are visualized to allow the engineer to select the prototype they want to further explore. Step 4: QD focuses on the user's selection to generate further solutions.}
	\label{fig:ideation_example}
\end{figure}

\noindent By applying prototype theory to the variety of designs produced by QD algorithms, \textit{computer-aided ideation} (CAI) is introduced (Fig.~\ref{fig:ideation_example}), allowing the same a posteriori articulation of preference as in the design by shopping paradigm. Although performance and diversity can both be formally described and optimized, design decisions are based on the intuition of the engineer, and cannot be automated. QD is used to discover a first set of optimal solutions. Then, by clustering similar solutions into classes and representative prototypes, the optimization process is guided by extracting seeds from the classes selected by the user, zooming in on a particular region in design space.
	
QD allows a paradigm shift in optimal engineering design, but integration of QD algorithms into the ideation process has yet to be studied extensively. In this work we introduce a CAI algorithm that takes advantage of recently introduced QD algorithms~\cite{Lehman2011,Mouret2015}. Prototype Discovery using Quality-Diversity (PRODUQD) \textipa{[pr@"d2kt]}, which performs a representative selection of designs, enables engineers to make design decisions more easily and influence the search for optimal solutions. PRODUQD can find solutions similar to a selection of prototypes that perform similarly well as solutions that were found by searching the entire design space with QD. By integrating QD algorithms and ideation a new framework for design is created; a paradigm which uses optimization tools to empower human intuition rather than replace it.

\section{Related Work}
\subsection{Quality-Diversity and Surrogate Assistance}

QD algorithms, like Novelty Search with Local Competition (NSLC)~\cite{Lehman2011} and Multidimensional Archive of Phenotypic Elites (MAP-Elites)~\cite{Mouret2015}, use a low-dimensional behavior or feature characterization, such as neural network complexity or curvature of a design, to determine similarity between solutions~\cite{Pugh2016}. Solutions compete locally in feature space, superseding similar yet less fit solutions. In MAP-Elites, the feature space consists of a discrete grid of behavior or feature dimensions, called a feature map. Every bin in the map is either empty or holds a solution, called an elite, that is currently the best performing one in its niche. QD is able to produce many solutions with a diverse set of behaviors and is very similar to the idea of Zwicky's morphological box~\cite{Zwicky1969} as it allows new creations by combining known solution configurations. QD algorithms perform many evaluations, making them unsuitable for design problems that need computationally expensive or real world evaluation.
	
To decrease the number of expensive objective evaluations, approximative surrogate models replace the objective function close to optimal solutions using appropriate examples~\cite{Jin2011}. To sample the design space effectively and efficiently, Bayesian Optimization is used. Given a prior over the objective function, evidence from known samples is used to select the next best observation. This decision is based on an acquisition function that balances exploration of the design space, sampling from unknown regions, and exploitation, choosing samples that are likely to perform well. This way, the surrogate model becomes more accurate in optimal regions during sampling. The most common surrogates used are Gaussian Process (GP) regression models~\cite{Rasmussen2006}. 

Surrogate assistance has been applied to QD with Surrogate-Assisted Illumination (SAIL)~\cite{Gaier2017}. In SAIL the GP model is pretrained with solutions evenly sampled in the parameter space with a Sobol sequence~\cite{Niederreiter1988}. The sequence allows iteratively finer sampling, approximating a uniform distribution. Then, using the upper confidence bound (UCB) acquisition function, an \textit{acquisition map} is created, containing a diverse set of candidate training samples.
UCB, described by the function $\mbox{UCB}(\textbf{x}) = \mu(\textbf{x}) + \kappa\sigma(\textbf{x})$, is a balance between exploitation ($\mu(\textbf{x})$, the mean prediction of the model), and exploration ($\sigma(\textbf{x})$, the model's uncertainty), parameterized by $\kappa$.

The acquisition map is first seeded with the previously acquired samples, assigning them to empty bins or replacing less performant solutions. MAP-Elites is then used in conjunction with the GP model to fill and optimize the acquisition map, using UCB as a fitness function and combining existing solutions from bins, illuminating the surrogate model through the "lens" of feature map. A candidate sample is created for every bin in the map. Then, the acquisition map is sampled using a Sobol sequence and selected solutions are evaluated using the expensive evaluation function. After gathering a given number of samples, the acquisition function is adapted by removing the model's uncertainty and the final prediction map, seeded with the set of known samples, is illuminated, producing a discrete map of diverse yet high-performing solutions.

\subsection{Dimensionality Reduction}
\label{sec:rel:DR}

Clustering depends on a notion of distance between points. The curse of dimensionality dictates that the relative difference of the distances of the closest and farthest data points goes to zero as the dimensionality increases~\cite{Beyer1999}. Clustering methods using a distance metric break down and cannot differentiate between points belonging to the same or to other clusters~\cite{Tomasev2016}. Dimensionality reduction (DR) methods are applied to deal with this problem. Data is often located at or close to a manifold of lower dimension than the original space. DR transforms the data into a lower-dimensional space, enabling the clustering method to better distinguish clusters~\cite{Tomasev2016}. Common DR methods are Principal Component Analysis (PCA) ~\cite{Pearson1901}, kernel PCA (kPCA)~\cite{Scholkopf1997}, Isomap~\cite{Tenenbaum2000}, Autoencoders~\cite{Hinton2006} and t-distributed Stochastic Neighbourhood Embedding (t-SNE)~\cite{Maaten2008}. 

t-SNE is commonly used for visualization and has been shown to be capable of retaining the local structure of the data, as well as revealing clusters at several scales. It does so by finding a lower-dimensional distribution of points $\mathbb{Q}$ that is similar to the original high-dimensional distribution $\mathbb{P}$. The similarity of datapoint $\boldsymbol{x}_j$ to datapoint $\boldsymbol{x}_i$ is the conditional probability ($p_{j|i}$ for $\mathbb{P}$ and $q_{j|i}$ for $\mathbb{Q}$, Eq. \ref{eq:pjk}), that $\boldsymbol{x}_i$ would pick $\boldsymbol{x}_j$ as its neighbor if neighbors were picked in proportion to their probability density under a Gaussian distribution centered at $\boldsymbol{x}_i$. The Student-t distribution is used to measure similarities between low-dimensional points $\boldsymbol{y}_i \in \mathbb{Q}$ in order to allow dissimilar objects to be modeled far apart in the map (Eq. \ref{eq:pjk}).

\begin{equation}
p_{j|i} = {e^{{-\left\lVert \boldsymbol{x}_i - \boldsymbol{x}_j \right\rVert^2} \over 2\sigma_i^2} \over \sum_{k \neq i} \bigg(e^{{-\left\lVert \boldsymbol{x}_i - \boldsymbol{x}_k \right\rVert^2} \over 2\sigma_i^2}\bigg)}
\text{, }q_{j|i} = {1 + \left\lVert \boldsymbol{y}_i - \boldsymbol{y}_j \right\rVert^2)^{-1} \over \sum_{k \neq i} (1 + \left\lVert \boldsymbol{y}_i - \boldsymbol{y}_k \right\rVert^2)^{-1})}
\label{eq:pjk}
\end{equation}

\noindent The local scale $\sigma_i$ is adapted to the density of the data (smaller in denser parts).
$\sigma_i$ is set such that \textit{perplexity} of the conditional distribution equals a predefined value. The perplexity of a distribution defines how many neighbors for each data point have a significant $p_{j|i}$ and can be calculated using the Shannon entropy $H(P_i)$ of the distribution $P_i$ around $x_i$ (Eq. \ref{eq:perp}).

\begin{multicols}{2}
	\noindent	
	\begin{equation}
	\text{Perp}(P_i) =  2^{-\sum_j p_{j|i} log_2 p_{j|i}}
	\label{eq:perp}
	\end{equation}
	\begin{equation}
	KL(\mathbb{P} \Vert \mathbb{Q})  = \sum_{i \ne j} p_{ij} log ({p_{ij} \over q_{ij}})
	\label{eq:KL}
	\end{equation}
\end{multicols}

\noindent Using the bisection method, $\sigma_i$ are changed such that $\text{Perp}(P_i)$ approximates the preset value (commonly 5-50). The similarity of $\boldsymbol{x}_j$ to $\boldsymbol{x}_i$ and $\boldsymbol{x}_i$ to $\boldsymbol{x}_j$ is absorbed with the joint probability $p_{ij}$. A low-dimensional map is learned that reflects all similarities $p_{ij}$ as well as possible. Locations $\boldsymbol{y}_i$ are determined by iteratively minimizing the Kullback-Leibler divergence of the distribution $\mathbb{Q}$ from the distribution $\mathbb{P}$ (Eq. \ref{eq:KL}) with gradient descent. 

\section{Prototype Discovery using Quality-Diversity}
\label{sec:automatedideation}

\begin{figure}
	\centering
	\includegraphics{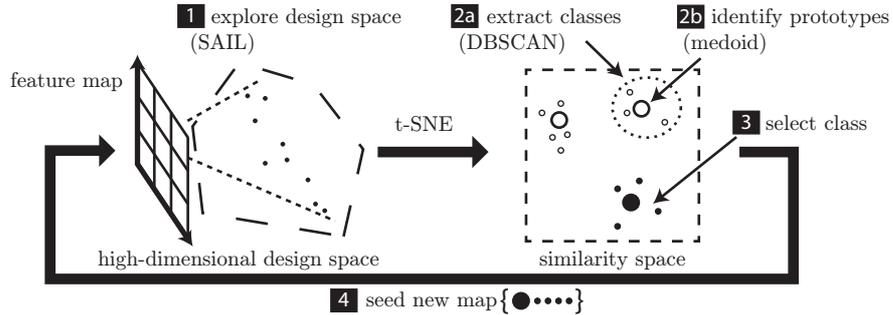}
	\caption{PRODUQD cycle - steps as in Figure \ref{fig:ideation_example}. Step 1: the design space is explored with the goal of filling the feature map. Step 2: (a) classes are extracted in a low-dimensional similarity space, and (b) prototypes are defined. Step 3: a selection is made. Step 4: seeds are extracted for the next iteration. }
	\label{fig:PRODUQD}
\end{figure}

PRODUQD is an example of a CAI algorithm (Fig. \ref{fig:PRODUQD}). Initially, the design space is explored with a QD algorithm. SAIL~\cite{Gaier2017} is used to explore the design space, creating high-performing examples of designs with varying features. These features can be directly extracted from design metrics, for instance the amount of head space in a car. SAIL produces a prediction map that contains a set of diverse high-performing solutions.

A similarity space is constructed using t-SNE. In this space, similar solutions are clustered into classes. Since no prior knowledge on the structure of optimality in design space is available and due to the stochastic nature of QD, the number and density of clusters is unknown. To group the designs into clusters we use the well-known density based clustering algorithm DBSCAN~\cite{Ester1996}\footnote{DBSCAN's parameterization is automated using the L Method~\cite{Salvador2003}.}.

A prototype is then extracted for every class. According to prototype theory~\cite{Rosch1975}, prototypes are those members of a class, ``with the most attributes in common with other members of the class and least attributes in common with other classes''. The most representative solution of a class is the member of a cluster that has the minimum distance to other members. The medoid is taken rather than a mean of the parameters, as this mean could lie in non-optimal or even invalid regions of the design space.

The prototypes are presented to the user, offering them a concise overview of the diversity in the generated designs. After the user selects one or more prototypes, the affiliated class members are used as seeds for the next SAIL iteration, serving as initial solutions in the acquisition and prediction maps. Initializing SAIL with individuals from the chosen class forces SAIL to start its search within the class boundaries. Using a subset of untested solutions of a particular class stands in contrast to SAIL, which focuses on searching the entire design space, seeding both maps with actual samples. Within each SAIL iteration, the GP surrogate is retrained whenever new solutions are evaluated. 

A precise definition of PRODUQD can be found in Alg. \ref{alg1}, including the use of the selected seeds $\mathcal{S}$ in SAIL. This ideation process explores the design space while taking into account on-line design decisions.

\begin{algorithm}
	\caption{Prototype Discovery using Quality-Diversity (PRODUQD)}
	\label{alg1}
	\begin{multicols}{2}
	\begin{algorithmic}
		\State $\mathcal{X} \gets Sobol_{1:G}$,
		$\mathcal{Y} \gets PE(\mathcal{X}), \mathcal{S}_0 \gets \mathcal{X}$ 
		\\\Comment{$PE$ = precise eval., $S_0$ = initial seed}
		\For {iter $ = 1 \to PE\_budget $}
		\State{\textbf{[1] Explore Design Space}}
		\State ($\mathcal{X}_{pred}, \mathcal{Y}_{pred}$) $=$ 
		\Call{SAIL}{$\mathcal{X},\mathcal{S}_{iter-1}$}
		\State{\textbf{[2] Extract Classes}}
		\State $\mathcal{X}_{red} =$ 
		\Call{T-SNE}{$\mathcal{X}$}
		\State $\mathcal{C} =$ 
		\Call{DBSCAN}{$\mathcal{X}_{red}$}
		\\\Comment{\textit{C} = class assignments}
		\State{\textbf{[3] Determine Prototypes}}
		
		\For {j $ = 1 \to |\mathcal{C}| $}
		\State $\mathcal{P} \gets$ 
		\Call{MEDOID}{$x_{red}, c_j$}
		\EndFor
		
		\State{\textbf{[4] Select Prototype(s)}}
		\State $p_{sel} =$ 
		\Call{SELECT}{$\mathcal{P}$}			
		\\\Comment{$p_{sel}$ = user selected prototype}
		\State{\textbf{[5] Extract Seeds}}
		\State $\mathcal{S} = \mathcal{X}, x \in c_{sel}$
		\\\Comment{$c_{sel}$ class belonging to $p_{sel}$}		
		\EndFor
		
		\\\Comment{Surrogate-Assisted Illumination}  
		\Function{SAIL}{$\mathcal{X},\mathcal{S}$}
		\Comment{samples, seeds}  
		\State{\textbf{[1] Produce Acquisition Map}}
		\For {iter $ = 1 \to PE\_budget $}
		\State $\mathcal{D} \gets (\mathcal{X}, \mathcal{Y})$ 
		\Comment{observation set}		
		\State $acq() \gets$ UCB($GP\_model(\mathcal{D})$)
		\State $(\mathcal{X}_{acq}, \mathcal{Y}_{acq}) =$ 
		\Call{MAP-E.}{$acq, \mathcal{S}$}\\\Comment{\textit{MAP-E. = MAP-Elites}}		
		\State $\mathbf{x} \gets \mathcal{X}_{acq}(Sobol_{iter})$
		\\\Comment{Select from acquisition map}
		\State $\mathcal{X} \gets \mathcal{X} \cup \mathbf{x}$,
		$\mathcal{S} \gets \mathcal{S} \cup \mathbf{x}$
		\State $\mathcal{Y} \gets \mathcal{Y} \cup PE(\mathbf{x})$
		\EndFor
		\State{\textbf{[2] Produce Prediction Map}}
		\State $\mathcal{D} \gets (\mathcal{X}, \mathcal{Y})$ 
		
		\State $\mathcal{GP} \gets GP\_model(\mathcal{D})$
		\State $pred() \gets mean(\mathcal{GP}(x))$
		
		\State $(\mathcal{X}_{pred}, \mathcal{Y}_{pred}) =$ 
		\Call{MAP-E.}{$pred, \mathcal{S}$}\\
		\Return ($\mathcal{X}_{pred}, \mathcal{Y}_{pred}$)
		\EndFunction
	\end{algorithmic}
\end{multicols}
\end{algorithm}

\section{Evaluation}
\label{sec:evaluation}

PRODUQD is a tool for CAI which allows the optimization and exploration of QD to be focused to produce designs which resemble user-chosen prototypes. We show that PRODUQD creates solutions of comparable performance to SAIL, produces models with the same level of accuracy, while directing the search towards design regions chosen by the user.

\subsubsection{2D Domain}
\begin{figure}
	\centering
	\includegraphics{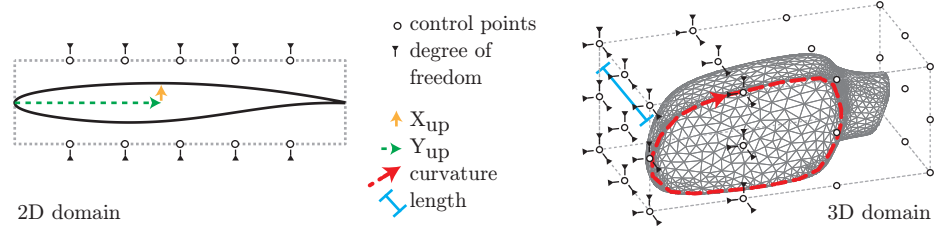}
	\caption{Left: 2D airfoil with control points and features ($X_{up}$, $Y_{up}$), right: control points of 3D mirror representation and features (curvature and length).}
	\label{fig:rep}
\end{figure}

PRODUQD and SAIL are applied to a classic design problem, similar to~\cite{Gaier2017}, but with a different representation. A high-performing 2D airfoil is optimized using free form deformation, with 10 degrees of freedom (Fig. \ref{fig:rep}). The base shape, an RAE2822 airfoil, is evaluated in XFOIL\footnote{\url{web.mit.edu/drela/Public/web/xfoil/}}, at an angle of attack of 2.7$\degree$ at Mach 0.5 and Reynolds number 10$^6$.

\noindent The objective is to find diverse deformations, minimizing the drag coefficient $c_D$ while keeping a similar lift force and area, described by $\mbox{fit}(x) = \mbox{drag}(x) \times p_{c_L}(x) \times p_A(x), \mbox{drag} = -log(c_D(x)), A = \mbox{area}$ and Eq. \ref{eq:fit1}-\ref{eq:fit2}. The feature map, consisting of the x and y coordinates of the highest point on the foil ($X_{up}$ and $Y_{up}$, see Fig. \ref{fig:rep}), is divided into a 25x25 grid.

\begin{multicols}{2}
	\noindent    
	\begin{equation}
	p_{c_L}(x)=\begin{cases}
	{c_L(x) \over c_{L_{0}}}^2, c_L(x)<c_{L_0}\\
	1\mbox{, otherwise} \enspace .
	\end{cases}
	\label{eq:fit1}
	\end{equation}
	\noindent
	\begin{equation}
	p_A(x) = \left( 1 - {|A-A_{0}| \over A_{0}} \right)^7
	\label{eq:fit2}
	\end{equation}
\end{multicols}

\subsubsection{3D Domain}
To showcase CAI on a more complex domain, the side mirror of the DrivAer~\cite{Heft2012} car model is optimized with a 51 parameter free form deformation (Fig. \ref{fig:rep} (right)). The objective is to find many diverse solutions while minimizing the drag force (in Newtons) of the mirror. The numerical solver OpenFOAM\footnote{\url{openfoam.org}, simulation at 11 m/s.} is used to determine flow characteristics and calculate the drag force. The feature map, consisting of the curvature of the edge of the reflective part of the mirror and the length of the mirror in flow direction, is divided into a 16x16 grid.

\subsubsection{Choice of Dimensionality Reduction Technique}
Various DR methods are analyzed as to whether they improve the clustering behavior of DBSCAN compared to applying clustering on the original dimensions. $\bar{G}_+$, a measure of the discordance between pairs of point distances and is robust w.r.t. differences in dimensionality~\cite{Tomasev2016}, is used as a metric. It indicates whether members of the same cluster are closer together than members of different clusters. A low value ($\geq 0$) indicates a high quality of clustering. PCA, kPCA, Isomap, t-SNE\footnote{Perplexity is set to 50, but at most half the number of samples.} and an Autoencoder are compared using DBSCAN on the latent spaces. t-SNE has been heavily tested for a dimensionality reduction to two dimensions. To allow a fair comparison, the same reduction was performed with the alternative algorithms.

\begin{table}
	\centering
	\caption{Quality of DR methods. Variance of the Autoencoder in parentheses.}
	\label{table:DR_quality1}
	\begin{tabular}{lllllll}
		\hline\noalign{\smallskip}
		& Original & PCA & kPCA & Isomap  & t-SNE & Autoencoder \\
		\noalign{\smallskip}
		\hline
		avg. $G_+$ score & 0.36 &  0.33  &  0.22 &   0.30 &   \textbf{0.05} &   0.454 (0.17)\\
		avg. number of clusters & 4 &  5 &   7  &   4  &  \textbf{10}  &   4 (1.23)\\
	\end{tabular}
\end{table}	
	
\noindent SAIL is performed 30 times on the 2D airfoil domain. For every run of SAIL, the dimensionality of the resulting predicted optima is reduced with the various methods and the optima are clustered with DBSCAN. Table \ref{table:DR_quality1} shows that t-SNE allows DBSCAN to perform about an order of magnitude better than using other methods. Although t-SNE is not a convex method, it shows no variance, indicating that the method is quite robust. The number of clusters found is about twice as high as using other methods, and since the cluster separation is of higher quality, t-SNE is selected as a DR method in the rest of this evaluation.

\subsubsection{Quantitative Analysis}
To show PRODUQD's ability to produce designs based on a chosen prototype, it is replicated five times, selecting a different class in each run. In every iteration of PRODUQD 10 iterations of SAIL are run to acquire 100 new airfoils. The first iteration starts with an initial set of 50 samples from a Sobol sequence. Then, the five classes containing the largest number of optima are selected, and the algorithm is continued in separate runs for each class. After each iteration, the we select the prototype that is closest to the one that was selected in the first iteration. PRODUQD runs are compared to the original SAIL algorithm, using the same number of samples, a total of 500.

\begin{figure}
	\centering
	\includegraphics{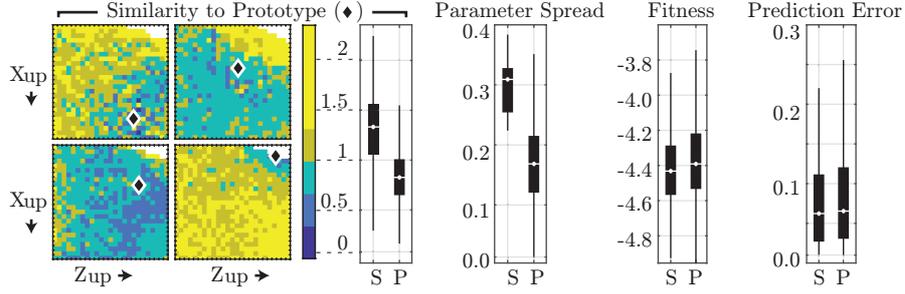}
	\caption{PRODUQD (P) produces designs that are more similar to the selected prototype than using SAIL (S), which is also visible in the smaller parameter spread. The produced designs have similar performance compared to SAIL's and the surrogate model is equally accurate. Left: final prototype similarity of four different PRODUQD prediction maps.}
\label{fig:evaluation:boxplots}
\end{figure}

\noindent The similarity of designs to prototypes of optima found in four separate runs, selecting a different prototype in each one, are shown in Fig.~\ref{fig:evaluation:boxplots} (left). The usage of seeds does not always prevent the ideation algorithm of finding optima outside of the selection, but PRODUQD produces solutions that are more similar to the selected prototype than SAIL. The parameter spread in solutions found with PRODUQD is lower than with SAIL. Yet the true fitness scores and surrogate model prediction errors of both SAIL and PRODUQD are very similar.

\begin{figure}
	\centering
	\includegraphics{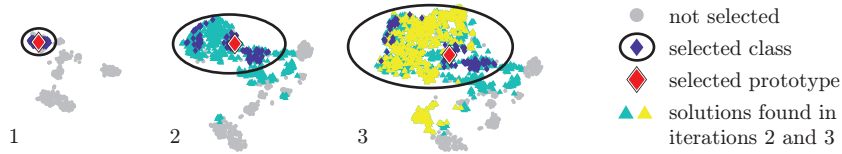}
	\caption{The region around the selected class is enlarged in similarity space and structure is discovered as more designs are added in later iterations. In each iteration the feature map is filled with solutions from the selected class.}
	\label{fig:evaluation:visual2D}
\end{figure}

Fig.~\ref{fig:evaluation:visual2D} shows the similarity space of three consecutive iterations. The effect of selection, zooming in on a particular region, can be seen by the fact that later iterations cover a larger part of similarity space, close to the prototype that was selected. Some designs still end up close to non-selected classes (in gray), which cannot be fully prevented without using constraints. PRODUQD is able to successfully illuminate local structure of the objective function around a prototype. It finds optima within a selected class of similar fitness to optima found in SAIL using no selection, and is able to represent the solutions in a class in a more concise way, using prototypes as representatives, shown by the decreased variance within classes (Fig.~\ref{fig:evaluation:boxplots}). 

The performance of PRODUQD's designs is comparable to SAIL while directing the search towards design regions chosen by the user.

\subsubsection{Qualitative Analysis}
\begin{figure}[t]
	\centering
	\includegraphics{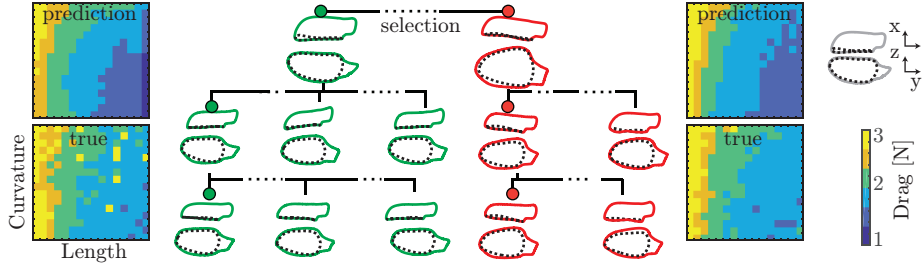}
	\caption{Phylogenetic tree of two PRODUQD runs diverging after first iteration, and predicted drag force maps (ground truth values are shown underneath).}
	\label{fig:evaluation:3D}
\end{figure}

A two-dimensional feature map, consisting of the curvature and the length of the mirror in flow direction (Fig. \ref{fig:rep}), is illuminated from an initial set of 100 car mirror designs from a Sobol sequence. After acquiring 200 new samples with SAIL, a prediction map is produced and from this set of solutions the two prototypes having the greatest distance to each other are selected and PRODUQD is continued in two separate instances, sampling another 100 examples. Then the newly discovered prototype that is closest to the one first selected is used to perform two more iterations, resulting in a surrogate model trained with 600 samples. The two resulting runs are shown in Fig.~\ref{fig:evaluation:3D}. Every branch in the phylogenetic tree of designs represents a selected prototype and every layer contains the prototypes found in an iteration. $18.8$ prototypes were found on average in each iteration. The surrogate model gives an accurate prediction of the drag force of all classes.

\section{Conclusion}
Quality-diversity algorithms can produce a large array of solutions, possibly impeding an engineer's capabilities of making a design decision. We introduce computer-aided ideation, using QD in conjunction with a state of the art dimensionality reduction and a standard clustering technique, grouping similar solutions into classes and their representative prototypes. These prototypes can be selected to constrain QD in a next iteration of design space exploration by seeding it with the selected class. A posteriori articulation of preference allows automated design exploration under the design by shopping paradigm. Decisions can be based on an engineer's personal experience and intuition or other ``softer'' design criteria that can not be easily formalized. PRODUQD, an example of such a CAI algorithm, allows an engineer to partially unfold a phylogenetic tree of designs by selecting prototypical solutions. 

The similarity space can be used continuously as it is decoupled from the feature map. This allows the diversity metric, the feature characterization, to change between iterations. The order in which the feature dimensions are chosen can be customized depending on the design process. For example, the engineer can start searching the design space in terms of diversity of design and later on switch to functional features. In future work, changes in the feature map and their effects on PRODUQD will be analyzed. Although seeding the map proves to be sufficient to guide QD towards the selected prototype, it is not sufficient to guarantee that QD only produces solutions within its class. Constraints could limit the search operation. Adding the distance to the selected prototype to the acquisition function could bias sampling to take place within the class. Finally, although the median solution might be most similar to all solutions within a class, one indeed might choose the fittest solution of a class as its representative.

CAI externalizes the creative design process, building up a design vocabulary by concisely describing many possible optimal designs with representative prototypes. Engineers can cooperate using this vocabulary to make design decisions, whereby ideation allows them to understand the design space not only in general, but around selected prototypes. CAI, a new engineering design paradigm, automates human-like search whilst putting the human back into the loop. 

\section*{Acknowledgments}
This work received funding from the German Federal Ministry of Education and Research (BMBF) under the Forschung an Fachhochschulen mit Unternehmen programme (grant agreement number 03FH012PX5 project "Aeromat"). The authors would like to thank Adam Gaier for their feedback.

\end{document}